**IEEE** *Access*

# Improved Fitness-Dependent Optimizer Algorithm

**Danial A. Muhammed[1], Soran AM. Saeed[2], Tarik A. Rashid[3] (IEEE Member)**

Danial Abdulkareem Muhammed, Computer Department, College of Science, University of Sulaimani, Sulaymaniyah, Iraq. danial.muhammed@univsul.edu.iq

[2]Soran AM. Saeed, Sulaimania Polytechnic University, Sulaymaniyah, KRG, Iraq. soran.saeed@spu.edu.iq

[3]Tarik A. Rashid, Computer Science and Engineering, University of Kurdistan Hewler, Erbil, KRG, Iraq. tarik.ahmed@ukh.edu.krd

**ABSTRACT** The fitness-dependent optimizer (FDO) algorithm was recently introduced in 2019. An improved FDO (IFDO) algorithm is presented in this work, and this algorithm contributes considerably to refining the ability of the original FDO to address complicated optimization problems. To improve the FDO, the IFDO calculates the alignment and cohesion and then uses these behaviors with the pace at which the FDO updates its position. Moreover, in determining the weights, the FDO uses the weight factor ($wf$), which is zero in most cases and one in only a few cases. Conversely, the IFDO performs $wf$ randomization in the [0-1] range and then minimizes the range when a better fitness weight value is achieved. In this work, the IFDO algorithm and its method of converging on the optimal solution are demonstrated. Additionally, 19 classical standard test function groups are utilized to test the IFDO, and then the FDO and three other well-known algorithms, namely, the particle swarm algorithm (PSO), dragonfly algorithm (DA), and genetic algorithm (GA), are selected to evaluate the IFDO results. Furthermore, the CECC06 2019 Competition, which is the set of IEEE Congress of Evolutionary Computation benchmark test functions, is utilized to test the IFDO, and then, the FDO and three recent algorithms, namely, the salp swarm algorithm (SSA), DA and whale optimization algorithm (WOA), are chosen to gauge the IFDO results. The results show that IFDO is practical in some cases, and its results are improved in most cases. Finally, to prove the practicability of the IFDO, it is used in real-world applications.

## I. INTRODUCTION

Since computers were developed, the focus has been on the aspects of probing unidentified solutions and searching for the best solution. Alan Turing utilized a search algorithm [1] in 1945 to break the enigma cipher of Germany during the Second World War. The advancement of practical methods and a dramatic rise in the volume of computation have caused difficulties in addressing real-life problems. Therefore, issues of quickly and proficiently solving complex problems via classic methods based on formal logic or mathematical programming have appeared [2]. Many algorithms have been created with a variety of methods to handle these constraints, and optimization problems are one of these methods. The optimization procedure obtains the best solution of a function by looking for a parameter. Existing solutions are denoted by sets of possible values, of which one is the best solution. Generally, solving optimization problems is the purpose of inventing optimization algorithms [3].

Based on the environment of the algorithms, there is a simple categorization of optimization algorithms that can separate them into two central groups: deterministic algorithms and stochastic algorithms. The first group, the deterministic algorithms, produces a similar set of answers when a similar preliminary starting point is used to begin the iterations; this is due to utilizing inclination, for instance, hill-climbing with a strict move sequence. Alternatively, the second groups, the stochastic algorithms, regularly produce different answers with similar preliminary values without utilizing inclination. On the other hand, there is a minor difference in the final values; a similar best solution would match the specified accuracy. Stochastic algorithms are categorized into two types: heuristic and metaheuristic [4].

Heuristic algorithms utilize trial and error to look for a solution; it is expected that they will take a feasible amount of time to achieve a solution. Likewise, heuristic algorithms tend to use different approaches in randomization techniques and local explorations [5]. Additional research and improvements on heuristic algorithms transformed them into metaheuristic algorithms, and these new groups of algorithms have superior performance compared to the heuristic algorithms; therefore, the prefix of "meta", which means "higher" or "beyond", was associated with them [6]. Nevertheless, these two terms (heuristic and meta-heuristic) are currently indistinguishable to scientists, although a slight dissimilarity exists in their meanings. Recently, meta-heuristic nature-inspired algorithms have been used professionally and effectively to solve recent nonlinear numerical global optimization difficulties. All meta-heuristic algorithms attempt to build some stability between local exploration and randomization [7].

Recently, existing real-world problems have become complicated, and considering space, time, and cost, it is impractical to explore all the conceivable solutions. Consequently, to solve such real-world problems, reasonable techniques that are low-cost and fast are essential. Hence, to determine how to address these difficulties, scientists have investigated natural occurrences and animal behaviors, for instance, how path selection occurs for ants, how evading the enemy and chasing prey occur for a group of birds, flies or fish, and how gravity works. Therefore, the name "nature-inspired algorithms" was selected for the algorithms that were inspired by nature [8]. There are many nature-inspired algorithms. The University of Michigan started to develop such algorithms in 1960 when Holland and his colleagues published a book about their GA and republished it in 1970 and 1983 [9]. Simulated annealing (SA) was implemented by Kirkpatrick et al. The motivation for the SA algorithm was the annealing process of metal [10].

PSO and ant colony optimization (ACO) are two commonly used swarm intelligence algorithms that were proposed by Kennedy and Eberhart in 1995 and Dorigo et al., 1996, respectively. PSO is inspired by the collective grouping behavior of birds in searching for food, and ACO is inspired by the nature of the ant, which has the ability to hold previous paths in its mind. [11-13]. The authors of the PSO thought these behaviors would help the optimization issues; then, other algorithms benefitted from the definitions used in the PSO algorithm. In the last two decades, various excellent intelligence swarms have been suggested, such as differential evolution (DE) in 1997, which was proposed by R. Storn and K. Price; it was a vector-based algorithm and performed better than GA in many applications [14].

In 2005, the artificial bee colony (ABC) algorithm was proposed by Karaboga and Basturk [15, 16]. Xin-She Yang created the firefly algorithm (FA) in 2009 [17], and then, the same year, CS was suggested by the same author [18]. Moreover, a bat-inspired algorithm was suggested by Xin-She in 2010 [19]. The artificial plant optimization algorithm (APOA) proposed by Bing Yu et al. in 2013 is inspired by the natural plant growing process. [20]. Additionally, in 2014, Li et al., offered a newly announced algorithm, animal migration optimization (AMO), which is inspired by swarm migration behavior in animals [21]. Later, Mirjalili A. S. proposed three algorithms: first, DA, in 2015, based on the behaviors related to attraction to food and evasion of enemies; second, WOA, in 2016; third, the salp swarm algorithm (SSA) in 2017 [22-24].

The novel ABC was altered with two modified ABCs created by Laizhong et al. In the first variant, an adaptive method for the population size (AMPS) was implemented by the authors [25], and in the second variant, the authors implemented a ranking-based adaptive ABC algorithm (ARABC) [26]; these variants were used for improvement exploitation in the original ABC algorithm. In 2019, Jaza Abdullah and Tarik Rashid developed a fitness-dependent optimizer or FDO algorithm. The FDO algorithm looks at the behaviors of bee swarms during reproduction and imitates swarm activities. Finding a different appropriate solution among various possible solutions forms a substantial part of this algorithm [27].

There are many other meta-heuristic optimization algorithms inspired by nature and utilized for difficult optimization problems, such as the evolutionary strategy (ES) [28], elephant herding optimization [29], fireworks algorithm (FWA) [30], biogeography-based optimization (BBO) [31], brain storm optimization [32], [33], earthworm optimization algorithm [34], krill herd algorithm (KH) [35-42], probability-based incremental learning (PBIL) [43], harmony search (HS) [44-46], bat algorithm (BA) [47, 48], monarch butterfly optimization (MBO) [49], and the moth search algorithm [50]. These algorithms cannot use all important information from instances in former iterations to direct their search in the present and future. Therefore, these algorithms can be divided into two groups. The first group, for instance, BBO [31, 50] and BA [47], is strictly independent of previous instances, and the second group, for instance, KH [35, 36], FWA [30, 51], and MBO [49] utilizes the instances that were best in earlier iterations [52].

Researchers have extensively utilized the abovementioned algorithms in many areas. However, there is no specific algorithm that achieves the most fitting solution for all optimization problems. Some algorithms yield better solutions for some specific problems than for others. Therefore, seeking adaptation in optimization techniques is an open problem [53].

In this paper, an improvement in fitness-dependent optimization (IFDO) has been developed from the FDO algorithm. In the FDO algorithm, the authors created the algorithm with a few characteristics of a scout. Jaza and Tarik described the main operator of the scout to update its location with its velocity (pace). Moreover, to manage weights, this operator typically relies on the fitness function value, and then, for the phases of exploitation and exploration, search agents are guided via these weights [27]. However, in IFDO, a scout exhibits other behaviors in addition to the pace, such as alignment and cohesion.

Moreover, the FDO, a weight factor ($wf$) was used to control the fitness weight. Nevertheless, the $wf$ was neglected in most cases [27]. However, in IFDO, the weight factor ($wf$) is used whenever a better fitness weight is obtained.

In the following, the papers' main contributions are briefly presented:

1) The IFDO algorithm is constructed by adding the behaviors of alignment and cohesion in updating the scout location and enhances the FDO algorithm in both the exploration and exploitation phases by considering reasonable covering of the search space to produce earlier convergence in the direction of global optimality.

2) The IFDO algorithm randomizes the $wf$ and utilizes it for each scout in each of the iterations.

3) One additional unique feature of IFDO is that when a better solution is obtained, a new $wf$ is generated in a new range to increase the chance of achieving the best solution in a shorter time (this is discussed further in section III).

The next sections describe this research. The second section presents the original algorithm FDO. The third section describes our improvements to the FDO algorithm. The fourth section shows the results and discussion; the performance information of the IFDO compared to the selected well-known and recent algorithms is specifically demonstrated, and then two real-world problems are addressed. The fifth section analyses the results and explains the role of the operators proposed in this study. Section 6 concludes the main points and suggests future research studies with the improved IFDO.

## II. FITNESS-DEPENDENT OPTIMIZER

The FDO can be divided into the scout bee searching process and the scout bee movement process. In the scout bee searching process, the algorithm makes the scout bees search for a suitable hive (solution) among many potential hives (solutions). Through the scout bee updating process, the algorithm utilizes a random walk and a fitness weight mechanism to move into a new position; accordingly, this section contains two parts.

### 1) Scout Bee Searching Process

The process of scout bees searching numerous possible hives to obtain a new proper hive means that the main part of this algorithm focuses on that process. In this algorithm, a proper solution is denoted by a scout bee that searches for a new hive. Moreover, meeting optimality means choosing the best hive among numerous hives. Furthermore, when the FDO begins execution, it defines an artificial scout population with random locations in an $Xi$ *(i=1, 2, …n)* space search by means of upper and lower boundaries. Through the execution, the FDO picks the global best solution. Finding a new hive (solution) in this algorithm is represented by a scout bee position. Scouts based on a random walk search in the search space for a more suitable solution; when the more suitable solution is revealed, the earlier solution is ignored. Nevertheless, if the scout cannot achieve a more suitable solution, then it uses the former solution with the expectation of finding a more suitable solution next time. Finally, in the case of not finding a more appropriate solution with the former solution, the scout will continue with the current solution, which is the best solution at that time.

### 2) Scout Bee Movement Process

In this algorithm, the scout, to obtain a better solution, updates its current position by adding pace. The updated artificial scout bee can be calculated according to equation (1) as follows:

$$X_{i,t+1} = X_{i,t} + pace \qquad (1)$$

where $i$ denotes the current search agent, $t$ denotes the current iteration, $X$ denotes an artificial scout bee (search agent), and pace denotes the movement rate and direction of the artificial scout bee. The pace is typically reliant on the fitness weight $fw$. Nevertheless, a random mechanism completely relies on the direction of the $pace$.

In FDO, the fitness weight $(fw)$ value is typically utilized to manage the $pace$. The algorithm determines the fitness weight $(fw)$ for every artificial scout using equation (2).

$$fw = \left| \frac{x^{*}_{i,t\ fitness}}{x_{i,t\ fitnees}} \right| - wf \qquad (2)$$

where $x^{*}_{i,t\ fitness}$ denotes the best global solution's fitness function value that has been revealed so far. $xi,t\ fitness$ denotes the current solution's value of the fitness function; $wf$ denotes a weight factor, randomly set between 0 and 1, which is used for controlling the $fw$.

Later, the algorithm considers some settings for $(fw)$, for instance, if $fw = 1$ or 0, and $x_{i,t\ fitnees} = 0$, the algorithm sets the pace randomly according to equation (3). On the other hand, if $fw > 0$ and $fw < 1$, then the algorithm generates a random number in the [-1, 1] range to make the scout search in every direction; when $r < 0$, pace is calculated according to equation (4), and when $r >= 1$, pace is calculated according to equation (5).

$$\begin{cases} fw = 1 \ or \ fw = 0 \ or \ x_{i,t\ fitness} = 0, \ \ pace = x_{i,t} * r \quad (3) \\ fw > 0 \ and \ fw < 1 \begin{cases} r < 0, pace = (x_{i,t} - x^{*}_{i,t}) * fw * -1 \ (4) \\ r \geq 0, \quad pace = (x_{i,t} - x^{*}_{i,t}) * fw \quad (5) \end{cases} \end{cases}$$

where $r$ denotes a random number in the range of [-1, 1], $xi,t$ denotes the current solution, and $x^{*}_{i,t}$ denotes the global best solution achieved thus far. Among various applications for random numbers, the FDO selects Levy flight because it considers further stable movement via its fair distribution curve [7].

The FDO pace is saved in every iteration for the accepted solution, and then it can be used next time.

## III. THE IMPROVED FITNESS-DEPENDENT OPTIMIZER

The IFDO is developed from the original FDO, which is an evolutionary optimization algorithm that was proposed by Jaza and Tarik [27]. The idea of this algorithm is essentially based on the generative process and collective decision-making used by bees. The bees search for many possible hives to obtain a new proper hive. Based on the original FDO, our proposed improved fitness-dependent optimizer is introduced, and it includes two phases: the updating of the scout bee position, which is improved by the functionalization of certain parameters, and the randomization of the weight factor $(wf)$ in the [0, 1] range. Accordingly, this section contains two parts.

### 1) Updating the Scout Bee Position

The IFDO, to create a different way of movement, applies order and cohesion, which are two vital signifiers of group motion; cohesion inside a group defines the distance between members, whereas members' alignment inside a group can be indicated by order when it is measured as divergence. Effective movement and maximization of the benefits of grouping for individual group members rely on better group cohesion and divergence [54].

In the original FDO, to achieve a more suitable solution, the scout bee adds pace to the current position in searching for new positions, as expressed in equation (1). In the IFDO, this equation is improved by adding two parameters, such as alignment and cohesion, to the pseudocode of the IFDO illustrated (see Figure (1)). In the following, the new movement of the artificial scout bee is expressed as:

$$X_{i,t+1} = X_{i,t} + pace + (alignment * {}^{1}/_{cohesion}) \qquad (6)$$

where $i$ is the current artificial scout bee (search agent), $t$ is the current iteration; the pace is the rate of the movement and the artificial bee direction, $X$ is an artificial bee, and alignment is the pace matching of scouts to that of other scouts in neighborhoods, and cohesion, is the inclination of scouts in the direction of the center of the mass of the neighborhood.

This improvement has been made in the light of scout bee behavior, which is always attracted to better solutions and avoids decreased chances of obtaining better solutions [27]. To calculate the alignment and cohesion behaviors, the scouts' neighbors' search landscape should be determined as shown in the pseudocode of the IFDO (see Figure (1)). In the IFDO, the search landscape of the artificial scout's neighbors is expressed as follows:

$$nl = \frac{lB}{2*PI} \qquad (7)$$

where $nl$ is the landscape of the neighbors, and $lB$ is the landscape boundary. To functionalize these two suggested parameters to update the scout bee position, it should be determined whether the scouts fall into the landscape of the neighbors $(ln)$, as shown in the pseudocode of the IFDO (see Figure (1)). The alignment and cohesion can be calculated according to equations (8) and (9).

$$\begin{cases} n = X - X_i, \ \ n = nl \ or \ n < nl \ , \ \ alignment_k = \frac{\sum_{k=1}^{N} pace_k}{N} \ (8) \\ n = X - X_i, \ \ n = nl \ or \ n < nl \ , \ \ cohesion_k = \frac{\sum_{k=1}^{N} X_k}{N} - X (9) \end{cases}$$

where $n$ represents a scout in the neighbors' landscape and the role of the variable $n$ is signifying which scout participates in determining the alignment and cohesion, $X$ represents the current scout's position, $N$ represents the neighborhood's number, $pace_k$ is the pace matching of scouts to that of other scouts in neighborhoods, and $x_k$ represents the position of the $k^{th}$ neighboring scout.

In the IFDO implementation, there are upper boundaries and lower boundaries for the dimensions of the agents to address weight values that are too large or small. See equations (10) and (11).

$$\begin{cases} wvb > ub, wvb = ub * nrd \qquad (10) \\ wvb < lb, wvb = lb * nrd \qquad (11) \end{cases}$$

where $wvb$ is the weight value of a bee, $ub$ is the upper boundary of the weight value of a bee, $nrd$ is the new random double value, and $lb$ is the lower boundary of the weight value of a bee.

The IFDO randomly moves the agents. The agent who remains still for finite time is the global best for this status; therefore, that agent randomly moves, and its movement will not be accepted until the agent obtains a better movement. See equation (3).

Because the FDO algorithm is PSO-based, this paper tries to add some PSO principles, such as alignment and cohesion, to improve the FDO algorithm from the perspective of convergence. Moreover, the IFDO has the same mathematical complexity as that of the FDO with a slight change in space complexity. The IFDO has time complexity $O (d*p + COF*p)$ for each iteration, where $d$ is the dimension of the problem, $p$ is the population size, and $COF$ is the cost of the objective function. On the other hand, IFDO has space complexity $O (COF*p + p*pace+(alignment*1/cohesion))$ for all iterations, where $pace+ (alignment*1/cohesion)$ is the best previous pace stored. Hence, for the total number of iterations, the time complexity in the IFDO is comparable. On the other hand, for the progress of iterations, its space complexity will be the same. Space complexity is slightly increased in the IFDO compared to the FDO due to the addition of two additional loops to calculate alignment and cohesion, although this increase is negligible, especially in modern computers, which have a substantial amount of memory space and computational time; this causes the IFDO to have decreased time complexity and better convergence.

### 2) Randomization Weight Factor

The original FDO uses pace as the degree of movement and the artificial bee direction. The regular fitness weight (fw) value is used to manage the pace. On the other hand, random mechanisms completely determine the pace direction. Hence, the minimization of fw is expressed according to equation (2).

The authors of the FDO algorithm stated that the weight factor is used to control the fitness weight and that the value of the weight factor is either 0 or 1; if $wf$ = 0, it is a more stable search, and if wf = 1, the convergence is high, and the chance of coverage is weak. Nevertheless, the authors mentioned that while the fitness function value entirely depends on the optimization problem, the reverse may also happen. Consequently, in our improved fitness-dependent optimizer, we use a random mechanism to control the fitness weight by generating a weight factor in the [0, 1] range, as shown in the pseudocode of the IFDO (see Figure (1)), to increase the IFDO performance, as is shown from the resulting test in section (4). In our proposed improvement, we change equation (2), as shown in equation (12).

$$fw = \left| \frac{x_{i,t \ fitness}^{*}}{x_{i,t \ fitnees}} \right| \qquad (12)$$

With equation (12), we find the fitness weight value and then check if it is less than or equal to the generated weight factor, as shown in the pseudocode of the IFDO (see Figure (1)); if it is, then the weight factor is ignored in controlling the fitness weight. Otherwise, the weight factor participates in controlling the fitness weight according to equation (13).

$$fw = fw - wf \quad (13)$$

This is a new way of finding the fitness weight, which is avoided by ignoring $wf$ in most cases, and $wf$ reasonably participates in many cases. In the IFDO, the weight factor is randomly set in every iteration for each scout, and a new $wf$ is generated in the new $[0, wf]$ range when a new, better solution is accepted, as shown in the pseudocode of the IFDO (see Figure (1)). From there, new $wf$ limited in $[0, wf]$ is better while for a new solution the IFDO will be more stable and higher coverage than the previous solution due to decreasing $wf$ for each iteration, as well as, it has more convergence than the setting $wf = 0$.

## IV. RESULTS AND DISCUSSION

This improved fitness-dependent optimizer's performance is verified using various standard test functions that exist in the literature; readers who are interested in knowing more about the methods of comparison can see references [27] [ 55] [57] [58]. Furthermore, the FDO implementation that can be found through the link https://github.com/Jaza-Abdullah/FDO-Java was downloaded; it was coded via the Java language. Then, the IFDO was created with the same language, and the IFDO algorithm was tested with the same parameter setting, the same test functions, and the same number of iterations as used in the FDO's tests. Moreover, the performance of the IFDO is evaluated against six state-of-the-art algorithms, namely, FDO, DA, GA, PSO, SSA, and WOA. The results of the tests of the 19 classical standard test functions and CEC-C06 tests for the different algorithms are taken from the original FDO work [27]. In addition, two real-world applications are optimized using the IFDO; therefore, this section consists of five parts, as follows:

### 1) Classical Benchmark Test Functions

The IFDO performance is tested with three groups of test functions [55]. There are various features for the test functions, such as unimodal, multimodal, and composite. To measure the algorithm's specific outcomes, these groups of tests are utilized. The stages of exploitation and convergence to infer a single optimum are verified by unimodal benchmark functions. On the other hand, there are many optimal solutions for the second feature (multimodal test functions); avoidance of local optima and stages of exploration are verified with this feature. It is worth mentioning that among the many optimal solutions, most are local optima, and there is only one global optimum. Avoiding local optimal solutions and moving toward a global optimum solution is essential to an algorithm. Additionally, with the third feature (composite test functions), various search areas can have various forms and large numbers of local optima. Composite test functions are generally moved, amalgamated, biased, and altered adaptations of other test functions. Difficulties that occur in real-world search areas can be identified by this type of standard function (see Tables 3, 4 and 5 in the appendix) [27].

*Initialize scout bee population $X_{t,i}$ (i = 1, 2, ..., n)*
*Generate random weight factor (wf) in [0, 1] range*
*while iteration (t) limit is not reached*
   *for each artificial scout bee $X_{t,i}$*
     *find best artificial scout bee $x_{t,i}^*$*
     *generate random-walk r in [-1, 1] range*
     *if( $X_{t,i}$ fitness == 0) (avoid dividing by zero)*
       *fitness weight = 0*
     *else*
       *calculate fitness weight, equation (12)*
       *if(fitness weight > wf)*
   *calculate fitness weight, equation (13)*
   *end if*
     *end if*
   *determine neighbors' search landscape (ln), equation (7)*
   *if(x-x,i < ln or x-x,i == ln)*
   *calculate alignment, equation (8)*
   *calculate cohesion, equation (9)*
   *end if*
    *if (fitness weight = 1 or fitness weight = 0)*
      *calculate pace using equation (3)*
    *else*
       *if (random number >= 0)*
         *calculate pace using equation (5)*
       *else*
         *calculate pace using equation (4)*
       *end if*
     *end if*
    *calculate $X_{t+1,i}$, equation (6)*
    *if( $X_{t+1,i}$ fitness < $X_{t,i}$ fitness)*
     *move accepted and pace saved*
   *generate new wf in [0, wf]*
    *else*
      *calculate $X_{t+1,i}$, equation (6)*
      *... with previous pace*
   *if ($X_{t+1,i}$ fitness < $X_{t,i}$ fitness)*
    *move accepted and save pace*
      *generate new wf in [0, wf]*
   *else*
      *maintain current position (don't move)*
    *end if*
    *end if*
   *end for*
*end while*

FIGURE 1. IFDO Pseudocode

To determine the average and standard deviation for each algorithm in Table (1) based on searching for the optimal solution, the algorithms in Table (1) are tested 30 times for 500 iterations and 30 scout bees each with 10 dimensions.
Parameter explanations for the DA, PSO, and the GA can be obtained in [55]. Moreover, there is only one parameter for the IFDO and the standard FDO, which is $wf$. For the FDO, in the test functions in Table (1), in only two of the cases (2 and 8), $wf$ is set to 1, and for all other cases, $wf$ is set to 0. In contrast, in our proposed algorithm (IFDO), $wf$ is set randomly in the [0, 1] range for all of the cases. However, this range will change when the algorithm detects a more suitable solution; for more detail, see Figure (1). During the test, only the test function TF8 is reduced to -2917375.29380209, and all of the other test functions are reduced to 0.0 (details of the conditions of the test functions can be found in Appendix Tables 3, 4 and 5). To confirm that the algorithm does not discriminate in the direction of origin, some degree of shifting is utilized for some of the test functions.

The IFDO results and the FDO, GA, DA, and PSO results are illustrated in Table (1). The results show that the IFDO in TF5, TF8, TF11, and TF12 was driven better overall in comparison with the selected comparator algorithms. However, the IFDO was worse than the other algorithms in TF6, TF7, and TF13. Moreover, the results of TF7, TF17, and TF18 showed that the IFDO was more comparable to the

The CEC global optimum is entirely bound to point 1 to be more appropriate. With the FDO, the three other recent algorithms for optimization, DA, WOA, and SSA, are tested for competitiveness with our proposed IFDO. Various motivations led to choosing these recent algorithms. First, the improved FDO, the original FDO, and the other chosen algorithms are all PSO-based algorithms. Second, in previous works, these algorithms were obviously used. Third, on both real-world problems and benchmark test functions, all of these algorithms have exceptionally good results. Fourth, the authors of these algorithms freely provided the algorithms'

| | TABLE 1 | | | | | | | | | |
| | FDO AND CHOSEN ALGORITHMS [27] WITH IFDO CLASSICAL BENCHMARK results | | | | | | | | | |
| Test Function | IFDO | | FDO | | DA | | PSO | | GA | |
| | AV. | ST. | AV. | ST. | AV. | ST. | AV. | ST. | AV. | ST. |
| TF1 | 5.38E-24 | 2.74E-23 | 7.47E-21 | 7.26E-19 | 2.85E-18 | 7.16E-18 | 4.2E-18 | 1.31E-17 | 748.5972 | 324.9262 |
| TF2 | 0.534345844 | 1.620259633 | 9.388E-6 | 6.90696E-6 | 1.49E-05 | 3.76E-05 | 0.003154 | 0.000811 | 5.971358 | 1.533102 |
| TF3 | 2.88E-07 | 6.90E-07 | 8.5522E-7 | 4.39552E-6 | 1.29E-06 | 2.1E-06 | 0.001891 | 0.003311 | 1949.003 | 994.2733 |
| TF4 | 2.60E-04 | 9.11E-04 | 6.688E-4 | 0.0024887 | 0.000988 | 0.002776 | 0.001748 | 0.002515 | 21.16304 | 2.605406 |
| TF5 | 1.94E+01 | 3.31E+01 | 23.50100 | 59.7883701 | 7.600558 | 6.786473 | 63.45331 | 80.12726 | 133307.1 | 85,007.62 |
| TF6 | 4.22E+06 | 8.15E-09 | 1.422E-18 | 4.7460E-18 | 4.17E-16 | 1.32E-15 | 4.36E-17 | 1.38E-16 | 563.8889 | 229.6997 |
| TF7 | 5.68E-01 | 3.14E-01 | 0.544401 | 0.3151575 | 0.010293 | 0.004691 | 0.005973 | 0.003583 | 0.166872 | 0.072571 |
| TF8 | -2.92E+06 | 2.24E+05 | -2285207 | 206684.91 | -2857.58 | 383.6466 | -7.1E+11 | 1.2E+12 | -3407.25 | 164.4776 |
| TF9 | 1.35E+01 | 6.66E+00 | 14.56544 | 5.202232 | 16.01883 | 9.479113 | 10.44724 | 7.879807 | 25.51886 | 6.66936 |
| TF10 | 5.18E-15 | 1.67E-15 | 3.996E-15 | 6.3773E-16 | 0.23103 | 0.487053 | 0.280137 | 0.601817 | 9.498785 | 1.271393 |
| TF11 | 0.525690405 | 8.90E-02 | 0.568776 | 0.1042672 | 0.193354 | 0.073495 | 0.083463 | 0.035067 | 7.719959 | 3.62607 |
| TF12 | 1.81E+01 | 2.57E+01 | 19.83835 | 26.374228 | 0.031101 | 0.098349 | 8.57E-11 | 2.71E-10 | 1858.502 | 5820.215 |
| TF13 | 4.10E+09 | 1.50E-05 | 10.2783 | 7.42028 | 0.002197 | 0.004633 | 0.002197 | 0.004633 | 68,047.23 | 87,736.76 |
| TF14 | 2.68E+07 | 4.68E-07 | 3.7870E-7 | 6.3193E-7 | 103.742 | 91.24364 | 150 | 135.4006 | 130.0991 | 21.32037 |
| TF15 | 4.03E-16 | 9.25E-16 | 0.001502 | 0.0012431 | 193.0171 | 80.6332 | 188.1951 | 157.2834 | 116.0554 | 19.19351 |
| TF16 | 9.14E-16 | 3.61E-16 | 0.006375 | 0.0105688 | 458.2962 | 165.3724 | 263.0948 | 187.1352 | 383.9184 | 36.60532 |
| TF17 | 2.38E+01 | 1.24E-01 | 23.82013 | 0.2149425 | 596.6629 | 171.0631 | 466.5429 | 180.9493 | 503.0485 | 35.79406 |
| TF18 | 2.24E+02 | 2.68E-05 | 222.9682 | 9.9625E-6 | 229.9515 | 184.6095 | 136.1759 | 160.0187 | 118.438 | 51.00183 |
| TF19 | 3.15E+01 | 1.32E-03 | 22.7801 | 0.0103584 | 679.588 | 199.4014 | 741.6341 | 206.7296 | 544.1018 | 13.30161 |

original FDO, whereas the results of TF10 and TF19 demonstrated that the IFDO outperformed the other competitor algorithms. Additionally, the results of TF1, TF3, TF4, TF9, TF14, TF15, and TF16, which are highlighted in green in Table (1), proved that the IFDO surpassed the original FDO, GA, PSO, and DA in all the situations.

### 2) CEC-C06 2019 Benchmark Test Functions

To further evaluate the IFDO, the algorithm was tested on 10 current test function sets of the CEC standard. Professor Suganthan and his colleagues enhanced these test functions for the optimization of a single objective problem [56]. A set of CEC standard test functions are planned to be used in the annual optimization competition "The 100-Digit Challenge", which is a common name for this set of test functions (see Table (2)). CEC01 to CEC03 are not similar to the test functions CEC04 to CEC10, while CEC01 to CEC03 are not shifted and rotated. However, a feature of scalability is utilized in both CEC01 to CEC03 and CEC04 to CEC10. Regarding the parameters, the CEC benchmark developer provided a set of parameters; the various dimensions for CEC01 to CEC03 are as shown in the Appendix in Table 6, and a 10-dimensional minimization problem in the [-100, 100] boundary range was set for the functions CEC04 to CEC10.

operating methods. It is worth mentioning that the parameter settings of the chosen algorithms were not changed during the test. The same settings were used for all the opponents, as shown in papers [27] [55] [57] [58]. Readers can access the MATLAB parameter setting arrangement and their implementations for the algorithms in this reference if desired [59]. Furthermore, the generated random weight factor (wf) in the [0, 1] range is used for all test functions; however, this $wf$ is regenerated in $[0, wf]$ for the next iteration if a better fitness weight ($fw$) is achieved (see the pseudocode in Figure (1)). To perform the test of IFDO and other competitors' algorithms as presented in Table (2), 30 agents with 500 iterations were applied to each algorithm.

In the cases of CEC02, CEC03, CEC09, and CEC10, the IFDO was equal to the original FDO; however, the standard deviation (SD) was changed somewhat. On the other hand, IFDO surpasses other competitors' algorithms in those cases. In cases CEC04 - CEC08, except for CEC06, the IFDO outperformed all of the opponents; however, in the case of CEC06, the IFDO performed worse than the DA, WOA, and

SSA but better than the original FDO. Finally, it is clear that the average IFDO, FDO, and WOA results are equal, whereas the standard deviation of WOA is equal to 0, which means there is no way to promote enhancement because similar results are obtained in all cases.

### 3) Quantitative Measurement Metrics

Two quantitative metrics were used for further investigation and detailed observation of IFDO, as shown in Figures 2 and 3. For each quantitative metric, among the unimodal standard functions TF1 to TF7, the first test function is chosen, among the multimodal standard test functions TF8 to TF13, the second test function is chosen, and among the composite standard functions TF14 to TF19, the third test function is chosen. For each investigation, searching the two-dimensional search space through 150 iterations was performed using 10 search agents.

of the convergence. During the test, the positions of the scout bees are logged from the start of the test to the end. Hence, this metric is simply a scout bee search history. At first, the whole area is rapidly discovered by the scout bee, and then, in the direction of optimality, they steadily move. Figure (2) presents the first quantitative metrics test.

The second measurement metric test illustrates the iteration process that measures the agent's global best convergence. When the number of iterations is increased, $x_i^*$ (the global best agent) is more precise, and when the scout bee focuses on the exploitation and local search, rapid changes are observed. See figure (3).

Generally, the IFDO has the ability to successfully explore

**TABLE 2**
**RESULTS OF THE IEEE ECE BENCHMARK 2019 [27]**

| Test Function | IFDO | | FDO | | DA | | WOA | | SSA | |
|---|---|---|---|---|---|---|---|---|---|---|
| | AV. | ST. | AV. | ST. | AV. | ST. | AV. | ST. | AV. | ST. |
| **CEC01** | 2651.198672 | 13944.10274 | 4585.27 | 20707.627 | $543 \times 10^8$ | $669 \times 10^8$ | $411 \times 10^8$ | $542 \times 10^8$ | $605 \times 10^7$ | $475 \times 10^7$ |
| **CEC02** | 4.000002146 | 1.00E-05 | 4.0 | 3.22414E-9 | 78.0368 | 87.7888 | 17.3495 | 0.0045 | 18.3434 | 0.0005 |
| **CEC03** | 13.70240422 | 4.82E-09 | 13.7024 | 1.6490E-11 | 13.7026 | 0.0007 | 13.7024 | 0.0 | 13.7025 | 0.0003 |
| **CEC04** | 31.19516293 | 12.91586061 | 34.0837 | 16.528865 | 344.3561 | 414.0982 | 394.6754 | 248.5627 | 41.6936 | 22.2191 |
| **CEC05** | 1.13187643 | 0.070551978 | 2.13924 | 0.085751 | 2.5572 | 0.3245 | 2.7342 | 0.2917 | 2.2084 | 0.1064 |
| **CEC06** | 12.12714515 | 0.52079368 | 12.1332 | 0.600237 | 9.8955 | 1.6404 | 10.7085 | 1.0325 | 6.0798 | 1.4873 |
| **CEC07** | 115.5677518 | 10.27465902 | 120.4858 | 13.59369 | 578.9531 | 329.3983 | 490.6843 | 194.8318 | 410.3964 | 290.5562 |
| **CEC08** | 4.940001939 | 0.891043403 | 6.1021 | 0.756997 | 6.8734 | 0.5015 | 6.909 | 0.4269 | 6.3723 | 0.5862 |
| **CEC09** | 2.0 | 3.10E-15 | 2.0 | 1.5916E-10 | 6.0467 | 2.871 | 5.9371 | 1.6566 | 3.6704 | 0.2362 |
| **CEC10** | 2.718281828 | 4.44E-16 | 2.7182 | 8.8817E-16 | 21.2604 | 0.1715 | 21.2761 | 0.1111 | 21.04 | 0.078 |

The first measurement metrics test demonstrates how the search space is covered by the scout bee and presents the course the search space, justifiably move in the direction of optimality and avoid local optima.

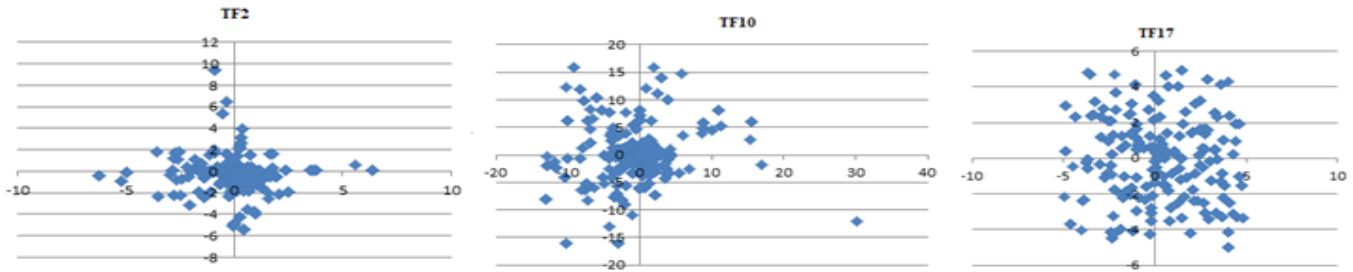

FIGURE 2. Using unimodal, multimodal, and composite test functions for the IFDO algorithm search history

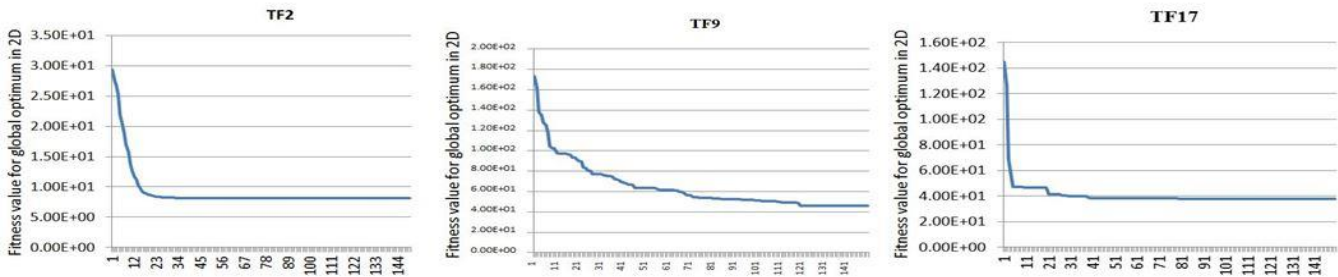

FIGURE 3. Using unimodal, multimodal, and composite test functions for the IFDO algorithm convergence curve

According to [60], if any algorithm's fitness value in minimization problems decreases with increasing iteration number, it reaches optimality.

### 4) Real World Applications of the IFDO

Real-world problems are solved via the IFDO and FDO; in this section, we performed two real-world applications.

The first application is the "aperiodic antenna array design," which was already tried by the original FDO. The second application is the "pedestrian evacuation model", which, to the best of our knowledge, is a new optimization problem that determines the best main door location inside an open area to evacuate people with greater efficiency. The results of the IFDO and FDO are evaluated for both real-world problems.

### A- USE OF THE IFDO ON APERIODIC ANTENNA ARRAY DESIGNS.

Developments in radio astronomy and radar methods from the 1960s drew significant attention to aperiodic antenna arrays. Thinned antenna arrays and non-uniform antenna arrays are shown in Figure (4).

Real-number vectors are needed to express a position in non-uniform arrays to optimize the element position with the intention of achieving the highest sidelobe level (SLL).

Additionally, as shown in equation (7), a confident boundary position of the element is needed to avoid discordant lobes. Interested readers can consult [61].

The 10 elements of a non-uniform isotropic array are shown in figure (5) and setting the outermost element to have an average element position of $d_{avg} = 0.5\lambda_0$ at position $2.25\lambda 0$ is a reason for optimizing the positions of the four elements alone. The limitations of this optimization problem with four dimensions are expressed in equation (14) as follows:

$$x\_i \in |x\_i - x\_j| (0,2.25) > \min\{x_i\} \ 0.25\lambda\_0 > 0.125\lambda\_0. \ i = 1,2,3,4. \ i \neq j. \qquad (14)$$

Nonetheless, there is no element that can be smaller than $0.125\lambda0$ or larger than $2.0\lambda0$. Due to these limitations, each element has a boundary between 0 and 2.25 because the element $2.25\lambda0$ is fixed, and the neighboring elements do not have the ability to be closer than $0.25\lambda0$. Equation (15) defines the problem of the fitness function:

$$f = max\{20 \ log \ |AF(\theta)| \ \} \qquad (15)$$

where

$$AF(\theta) = \sum_{i=1}^{4} \cos[(\cos\theta - \cos\theta_s)2\pi x_i] + \cos[(\cos\theta - \cos\theta_s)2.25 \times 2\pi] \qquad (16)$$

For this work, Figure (5) shows that $\theta_s = 90°$ [62].

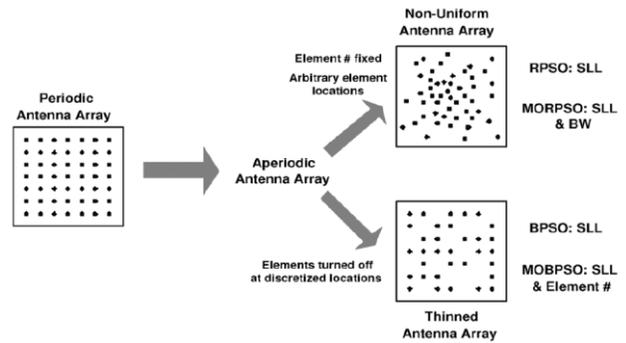

FIGURE 4. A thinned antenna array and a non-uniform antenna array [61].

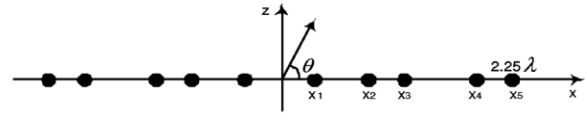

FIGURE 5. Ten-element arrangements in the array [62].

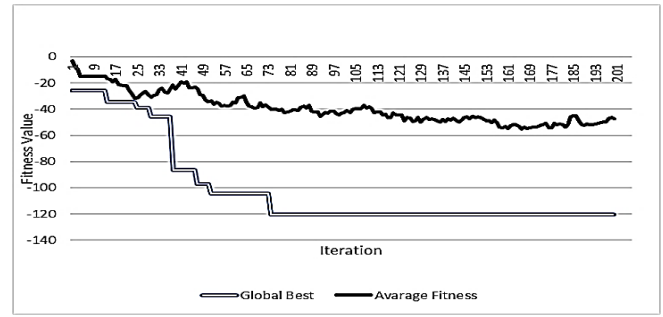

FIGURE 6. The average fitness and global optimum as a result of optimizing aperiodic antenna array designs in 200 iterations with 20 artificial scout bees using the standard FDO.

Based on the limitations stated in equation (14), for twenty artificial scout bees within 200 iterations, the original FDO algorithm was utilized to optimize this problem. Moreover, based on equation (15), the average fitness value and the global best fitness in each iteration are shown in Figure (6). The results indicate that with the element locations {0.713,1.595,0.433,0.130} in iteration 78, the global best solution was achieved.

Likewise, regarding the mentioned restrictions of this problem, similar to the original FDO, this problem was optimized using the IFDO algorithm in 200 iterations for twenty search agents (artificial bees), as shown in Figure (7), based on equation (15), which contains the average fitness value and the global best fitness in each iteration. The result shows that with element locations {0.701, 1.552, 0.402, 0.103}, the global best solution was achieved in iteration 29. Consequently, from both the IFDO and FDO results, it clearly appears that the IFDO is better for optimizing this problem due to its increasing capability of making better decisions in exploring better hives among the existing potential hives by adding alignment and cohesion when the scout wants to go to a different location in the defined space search; it also avoids unsuitable exploitation in achieving a better solution when, for every achieved better solution, the IFDO generates a new $wf$ to control the $fw$ (see the pseudocode in Figure (1)).

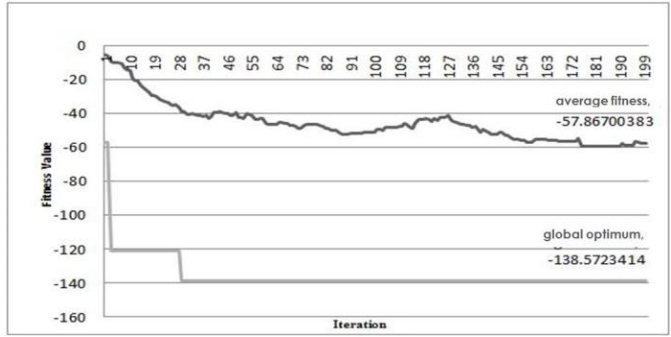

FIGURE 7. The average fitness and global optimum as a result of optimizing aperiodic antenna array designs in 200 iterations with 20 artificial scout bees using the IFDO.

### B-IFDO VS THE FDO ON A PEDESTRIAN EVACUATION MODEL.

In the last two decades, scenarios involving the evacuation of crowds and pedestrians have been studied in many works to reduce the negative aspects of emergency situations, such as deaths, damages, and injuries [63]. In this part of this paper, we create a simple pedestrian evacuation model based on a cellular automata model (see Figure (8)), fuzzy logic ideas, and statistical equations. Readers who desire to know how this evacuation model is created and how the ideas of fuzzy logics and statistical equations are utilized to define the pedestrians' desired speeds can access reference [64]. Additionally, the evacuation time of each pedestrian is calculated via the pedestrian's desired speed and its distance from the exit door as expressed in equation (17), and the average of the evacuation time of the pedestrians is used as the average fitness value.

$$evacTime = (dist/2) * desiredSpeed \quad (17)$$

where $dist$ represents the pedestrian's distance from the door exit locations, which is calculated from the equation of distance (18), and $desiredSpeed$ represents the pedestrian's speed.

$$dist = \sqrt{(x_2 - x_1)^2 + (y_2 - y_1)^2} \quad (18)$$

where $x_2$ and $y_2$ represent the coordinates of the exit door location, and $x_1$ and $y_1$ represent the coordinates of the pedestrian's location.

Finally, both the IFDO and FDO algorithms are applied to this model to achieve the global best solution by finding the best location of the main door through which to evacuate people during the evacuation process. The results showed that the IFDO was more efficient and reached the optimum solution with only 38 iterations, whereas the FDO reached the optimum solution with 57 iterations. Figure (9) shows the results of both algorithms.

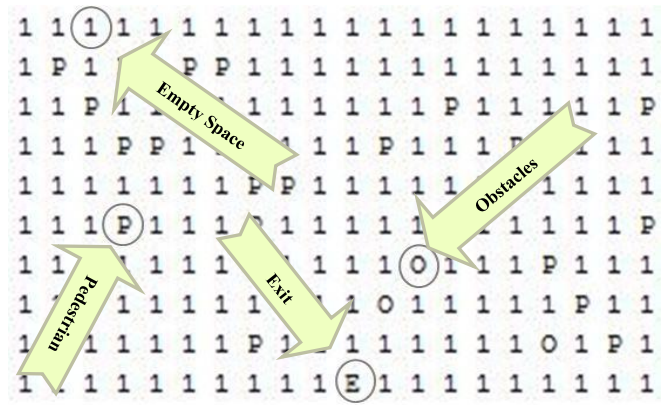

FIGURE 8. The area of the pedestrian evacuation model.

The reasons behind the IFDO's efficiency are related to the selected parameters, alignment, and cohesion, in updating the position of the artificial scout bees, which makes the algorithm perform better explorations in finding a suitable solution in the landscape. Second, the randomization in defining $wf$ in every iteration for each scout bee when a better solution is achieved makes the algorithm avoid unnecessary exploitations to gain a better solution. Third, the IFDO, as regards covering a reasonable search space, converges sooner to global optimality.

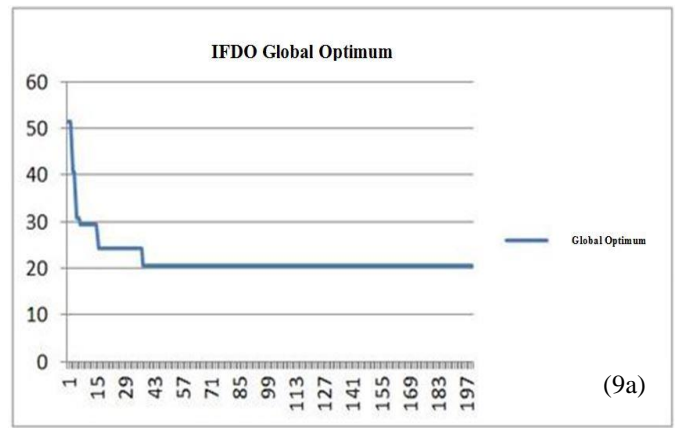

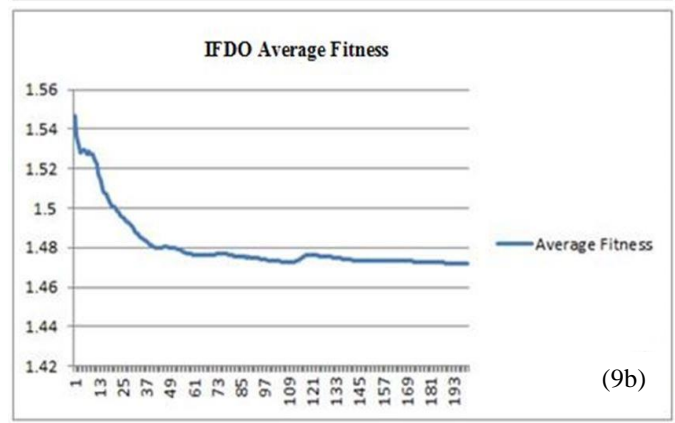

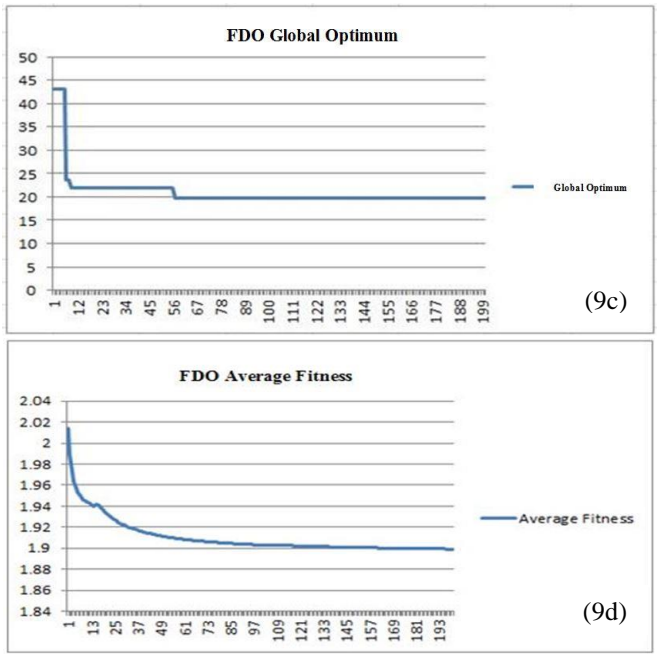

FIGURE 9. IFDO and FDO global optimum and average fitness
(a) IFDO global optimum, (b) IFDO average fitness, (c) FDO global optimum, and (d) FDO average fitness

For both FDO and IFDO, after testing on various real-world applications and classical and modern benchmark test functions, it was found that their performance depended on the number of search agents. Hence, both algorithms are limited to using a small number of search agents; for example, the accuracy of the algorithms suffers noticeably when they use fewer than five search agents. Conversely, using a large number of search agents enhances the accuracy and rate with more space and time.

### 5) IFDO VS FDO Execution Time

Here, execution time is considered for various tests, such as classical benchmark test functions, modern IEEE CEC 2019 benchmark test functions, and two real-world applications, aperiodic antenna array designs (AAAD) and pedestrian evacuation models (PEMs). The results of the total time are briefly provided in Tables 7, 8, and 9.

**TABLE 7**
**RESULTS OF THE IFDO VS FDO EXECUTION TIME FOR THE IEEE ECE BENCHMARK 2019**

| Test function | Execution time | |
|---|---|---|
| | IFDO | FDO |
| CEC01 | 15 seconds | 2 seconds |
| CEC02 | 19 seconds | 20 seconds |
| CEC03 | 21 seconds | 6 seconds |
| CEC04 | 16 seconds | 16 seconds |
| CEC05 | 14 seconds | 20 seconds |
| CEC06 | 49 minutes 46 seconds | 45 minutes 38 seconds |
| CEC07 | 21 seconds | 55 seconds |
| CEC08 | 31 seconds | 32 seconds |
| CEC09 | 34 seconds | 34 seconds |
| CEC10 | 29 seconds | 35 seconds |

**TABLE 8**
**RESULTS OF THE IFDO VS FDO EXECUTION TIME FOR THE CLASSICAL BENCHMARK TEST FUNCTIONS**

| Test function | Execution time | |
|---|---|---|
| | IFDO | FDO |
| TF1 | 15 seconds | 16 seconds |
| TF2 | 19 seconds | 17 seconds |
| TF3 | 21 seconds | 18 seconds |
| TF4 | 16 seconds | 16 seconds |
| TF5 | 14 seconds | 17 seconds |
| TF6 | 23 seconds | 18 seconds |
| TF7 | 21 seconds | 24 seconds |
| TF8 | 32 seconds | 31 seconds |
| TF9 | 34 seconds | 32 seconds |
| TF10 | 29 seconds | 33 seconds |
| TF11 | 40 seconds | 35 seconds |
| TF12 | 41 seconds | 44 seconds |
| TF13 | 55 seconds | 58 seconds |
| TF14 | 3 seconds | 3 seconds |
| TF15 | 7 seconds | 9 seconds |
| TF16 | 29 seconds | 27 seconds |
| TF17 | 25 seconds | 22 seconds |
| TF18 | 26 seconds | 27 seconds |
| TF19 | 20 seconds | 24 seconds |

**TABLE 9**
**RESULTS OF THE IFDO VS FDO EXECUTION TIME FOR REAL-WORLD APPLICATIONS**

| Application | Execution time | |
|---|---|---|
| | IFDO | FDO |
| AAAD | 47 seconds | 44 seconds |
| PEM | 28 seconds | 31 seconds |

From the results shown in table (7), the execution times of the modern IEEE CEC 2019 benchmark test functions for both the IFDO and FDO were relatively the same; for instance, the IFDO had a smaller total time of execution than the FDO in the execution of the CEC02, CEC05, CEC07, CEC08, and CEC10 cases; however, the IFDO took a larger portion of the total time to execute the CEC01, CEC03, and CEC06 cases. The IFDO and FDO took the same total time to execute the CEC04 and CEC09 cases. Moreover, the results of the classical benchmark test functions in table (8) show that the IFDO requires less time than the FDO to execute most of the test functions, such as TF1, TF5, TF7, TF10, TF12, TF13, TF15, TF18, and TF19, the same amount of time in a few cases, such as TF4 and TF14, and more time in some cases, such as TF2, TF3, TF6, TF8, TF9, TF11, TF16, and TF17. Finally, the execution time results of the real-world applications in Table (9) illustrate that the IFDO is more capable than the FDO from the perspective of spending time on PEM real-world applications, whereas it is not as powerful as the FDO in executing the AAAD.

## V. ANALYSIS OF THE RESULTS

The IFDO modified the FDO in both scout bee movements, to update their positions, and weight factor ($wf$), to control the fitness weight ($fw$), to find a better solution. From the results and discussion, it appears that these changes improve both exploration and exploitation. From there, these changes improve the time complexity and convergence. To evaluate this idea, readers can reference subsection IV in subsections 1, 2, 3, 4, and 5 to see that after the modifications, the IFDO was better in the classical benchmark test function results than the other competing algorithms. For instance, in TF1, TF3, TF4, TF9,

TF14, TF15, TF16, as well as in TF7, TF17, and TF18, the results showed the IFDO was more similar to the original FDO, while the results of TF10 and TF19 confirmed that the IFDO outperformed the other competing algorithms. Moreover, the IFDO had better results in cases CEC04-CEC08 than the opponents, except for case CEC06, in which it had a worse result than the opponents and a better result than the FDO. On the other hand, in cases CEC02, CEC03, CEC09, and CEC10, although the standard deviation was different from that in the original FDO, the IFDO was equivalent to that in the original FDO. Furthermore, the results of the quantitative measurement metrics revealed that the IFDO had the ability to successfully explore the search space, move toward optimality, and avoid the local optima. Additionally, the IFDO and FDO were used with real-world applications in 200 iterations for twenty search agents (artificial bees), and the IFDO outperformed the FDO algorithm. For example, in the aperiodic antenna array designs, the IFDO reached optimality with just 29 iterations, while the number of iterations needed in the FDO was 78. In the pedestrian evacuation model, the IFDO reached optimality in only 38 iterations, while the FDO required 57 iterations. From these results, it is possible to say that IFDO generally had a better performance in reaching optimality and better exploration and exploitation. Finally, the IFDO was compared with the FDO from the perspective of execution time. For this purpose, the classical benchmark test functions, IEEE CEC 2019 benchmark test functions, and two real-world applications, AAAD and PEM, were utilized. From the results, both IFDO and FDO were relatively similar in most of the results for the classical benchmark and the IEEE CEC 2019 benchmark test functions. However, the results of these algorithms in optimizing the two real-world applications were generally different. For instance, the IFDO required a shorter time than the FDO to optimize PEM: 28 seconds and 31 seconds, respectively. Conversely, the IFDO required a larger portion of time than the FDO to optimize AAAD: 47 seconds and 44 seconds, respectively.

## VI. CONCLUSION

Improvements have been made to the fitness-dependent optimizer from two main perspectives. First, for updating the artificial scout bee position, in the IFDO, two additional parameters were added to the position update equation in the original FDO: alignment and cohesion. Second, the weight factor ($wf$) was changed from a stable value to a random value in controlling the fitness weight of the FDO algorithm. These changes were made in the IFDO with the aim of moving the scout bees toward optimality with better performance. To evaluate the performance of the IFDO, it was tested with 19 single-objective benchmark test functions (unimodal, multimodal and composite test functions). Moreover, the 10 modern benchmarks of CEC-C06 were utilized to test the IFDO. Furthermore, quantitative measurement metrics were used to show that the IFDO succeeded in exploring the search space, moving towards optimality, and avoiding the local optima. Additionally, both the IFDO and FDO were used to execute the classical benchmark test functions, IEEE CEC 2019 test functions, and two real-world applications. Each test function's total time of execution was specified and compared.

The results of the IFDO tests with the classic and modern test functions were compared to those of the FDO, two other distinguished algorithms (GA and PSO), and three state-of-the-art algorithms (SSA, WOA, and DA). According to the results, the IFDO, except for some cases in which it had comparable results, outperformed the preferred algorithms in most cases. It could be said that this advancement was due to the modification in updating the artificial scout position, which led to more convenient exploration during the search for a better solution among many potential hives (solutions), and due to the randomization of the $wf$ for each scout bee in every iteration, which led to a better $fw$ participating in making better decisions in the exploitation to find better solutions. Additionally, the IFDO produced faster convergence to global optimality when considering rational coverage of the search space. On the other hand, the use of various numbers of scout bees affected the accuracy, cost, time, and space of the algorithm. When more than five scout bees were used, the enhanced accuracy of the algorithm could be clearly seen; however, a smaller number of scout bees led to decreased accuracy of the algorithm. In addition, to confirm that the IFDO has the ability to address real-life applications, two real-world problems were selected: the first problem was an existing real-world "aperiodic antenna array design" problem, and the second problem was a real-world crowd evacuation problem that we created. In both applications, the IFDO outperformed the original FDO; in the first application, the FDO needed 78 iterations to discover the global optimal solution, whereas the IFDO needed only 29 iterations to obtain the global optimal solution. Additionally, in the second application, the IFDO outperformed the original FDO; although the IFDO needed only 38 iterations to obtain the optimal global solution, the FDO needed 57 iterations to achieve the same result. It is worth mentioning that because this performance is an improvement compared with the original FDO, the improved fitness-dependent optimizer was selected as the official name of this improved algorithm. This proposed algorithm is more suitable for application fields of engineering, design, industry, helath, education, energy, and evacuation.

In future studies, multiobjective and binary objective optimization problems will be tested with the IFDO. Finally, adaptation and hybridization of the IFDO with other algorithms will be the main focus of future work. Also, the performance of IFDO can be further evaluated against other popular algorithms, such as WOA-BAT [65], Donkey and Smuggler Optimisation [66], and Modified Grey Wolf Optimiser [67], Modifications of Dragonfly Algorithm [68], Modifications of Backtracking Algorithm [69].

## VIII. APPENDIX

TABLE 3
Unimodal standard functions [**30**].

| Functions | Dimension | Range | Shift position | $f_{min}$ |
|---|---|---|---|---|
| $TF1(x) = \sum_{i=1}^{n} x_i^2$ | 10 | [-100, 100] | [-30, -30, ... -30] | 0 |
| $TF2(x) = \sum_{i=1}^{n} \|x_i\| + \prod_{i=1}^{n} \|x_i\|$ | 10 | [-10,10] | [-3, -3, ... -3] | 0 |
| $TF3(x) = \sum_{i=1}^{n} \left( \sum_{j-1}^{i} x_j \right)^2$ | 10 | [-100, 100] | [-30, -30, ... -30] | 0 |
| $TF4(x) = \max_i \{\|x\|, 1 \leq i \leq n\}$ | 10 | [-100, 100] | [-30, -30, ... -30] | 0 |
| $TF5(x) = \sum_{i=1}^{n-1} [100(x_{i+1} - x_i^2)^2 + (x_i - 1)^2]$ | 10 | [-30,30] | [-15, -15, ... -15] | 0 |
| $TF6(x) = \sum_{i=1}^{n} ([x_i + 0.5])^2$ | 10 | [-100, 100] | [-750, ... -750] | 0 |
| $TF7(x) = \sum_{i=1}^{n} i x_i^4 + random[0,1]$ | 10 | [-1.28,1.28] | [-0.25, ...-0.25] | 0 |

TABLE 4
(10 DIMENSIONAL) MULTIMODAL STANDARD FUNCTIONS [30].

| Functions | Range | Shift position | $f_{min}$ |
|---|---|---|---|
| $TF8(x) = \sum_{i=1}^{n} -x_i^2 \sin\left(\sqrt{\|x_i\|}\right)$ | [-500, 500] | [-300, ... -300] | -418.9829 |
| $TF9(x) = \sum_{i=1}^{n} [x_i^2 - 10\cos(2\pi x_i) + 10]$ | [-5.12,5.12] | [-2, -2, ...-2] | 0 |
| $TF10(x) = -20exp\left(-0.2\sqrt{\sum_{i=1}^{n} x_i^2}\right) - exp\left(\frac{1}{n}\sum_{i=1}^{n}\cos(2\pi x_i)\right) + 20 + e$ | [-32, 32] | | 0 |
| $TF11(x) = \frac{1}{4000}\sum_{i=1}^{n} x_i^2 - \prod_{i=1}^{n}\cos\left(\frac{x_i}{\sqrt{i}}\right) + 1$ | [-600, 600] | [-400, ... -400] | 0 |
| $TF12(x) = \frac{\pi}{n}\{10 \sin(\pi y_1) + \sum_{i=1}^{n-1}(y_i - 1)^2[1 + 10\ sin^2(\pi y_{i+1})] + (y_n - 1)^2\} + \sum_{i=1}^{n} u(x_i, 10, 100, 4).$ $y_i = 1 + \frac{x+1}{4}.$ $u(x_i, a, k, m) = \begin{cases} k(x_i - a)^m & x_i > a \\ 0 & -a < x_i < a \\ k(-x_i - a)^m & x_i < -a \end{cases}$ | [-50,50] | [-30, 30, ... 30] | 0 |
| $TF13(x) = 0.1\{sin^2(3\pi x1) + \sum_{i=1}^{n}(x_i - 1)^2[1 + sin^2(3\pi x_i + 1)] + (x_n - 1)^2[1 + sin^2(2\pi x_n)]\} + \sum_{i=1}^{n} u(x_i, 5, 100, 4).$ | [-50,50] | [-100, ... -100] | 0 |

TABLE 5
COMPOSITE STANADRD FUNCTIONS [30].

| Functions | Dimension | Range | $f_{min}$ |
|---|---|---|---|
| **TF14 (CF1)**<br>$f1, f2, f3 \dots f10 =$ Sphere function<br>$\delta1, \delta2, \delta3 \dots \delta10 = [1,1,1, \dots 1]$<br>$\lambda1, \lambda2, \lambda3 \dots \lambda10 = \left[\dfrac{5}{100}, \dfrac{5}{100}, \dfrac{5}{100}, \dots \dfrac{5}{100}\right]$ | 10 | [-5, 5] | 0 |
| **TF15 (CF2)**<br>$f1, f2, f3 \dots f10 =$ Griewank's function<br>$\delta1, \delta2, \delta3 \dots \delta10 = [1,1,1, \dots 1]$<br>$\lambda1, \lambda2, \lambda3 \dots \lambda10 = \left[\dfrac{5}{100}, \dfrac{5}{100}, \dfrac{5}{100}, \dots \dfrac{5}{100}\right]$ | 10 | [-5, 5] | 0 |
| **TF16 (CF3)**<br>$f1, f2, f3 \dots f10 =$ Griewank's function<br>$\delta1, \delta2, \delta3 \dots \delta10 = [1,1,1, \dots 1]$<br>$\lambda1, \lambda2, \lambda3 \dots \lambda10 = [1,1,1, \dots 1]$ | 10 | [-5, 5] | 0 |
| **TF17 (CF4)**<br>$f1, f2 =$ Ackley's function<br>$f3, f4 =$ Rastrigin's function<br>$f5, f6 =$ Weierstrass function<br>$f7, f8 =$ Griewank's function<br>$f9, f10 =$ Sphere function<br>$\delta1, \delta2, \delta3 \dots \delta10 = [1,1,1, \dots 1]$<br>$\lambda1, \lambda2, \lambda3 \dots \lambda10 = \left[\dfrac{5}{32}, \dfrac{5}{32}, 1,1, \dfrac{5}{0.5}, \dfrac{5}{0.5}, \dfrac{5}{100}, \dfrac{5}{100}, \dfrac{5}{100}, \dfrac{5}{100}\right]$ | 10 | [-5, 5] | 0 |
| **TF18 (CF5)**<br>$f1, f2 =$ Rastrigin's function<br>$f3, f4 =$ Weierstrass function<br>$f5, f6 =$ Griewank's function<br>$f7, f8 =$ Ackley's function<br>$f9, f10 =$ Sphere function<br>$\delta1, \delta2, \delta3 \dots \delta10 = [1,1,1, \dots 1]$<br>$\lambda1, \lambda2, \lambda3 \dots \lambda10 = \left[\dfrac{1}{5}, \dfrac{1}{5}, \dfrac{5}{0.5}, \dfrac{5}{0.5}, \dfrac{5}{100}, \dfrac{5}{100}, \dfrac{5}{32}, \dfrac{5}{32}, \dfrac{5}{100}, \dfrac{5}{100}\right]$ | 10 | [-5, 5] | 0 |
| **TF19 (CF6)**<br>$f1, f2 =$ Rastrigin's function<br>$f3, f4 =$ Weierstrass function<br>$f5, f6 =$ Griewank's function<br>$f7, f8 =$ Ackley's function<br>$f9, f10 =$ Sphere function<br>$\delta1, \delta2, \delta3 \dots \delta10 = [0.1, 0.2, 0.3, \ 0.4, 0.5, 0.6, 0.7, 0.8, 0.9, 1]$<br>$\lambda1, \lambda2, \lambda3 \dots \lambda10 = \left[0.1 * \dfrac{1}{5}, 0.2 * \dfrac{1}{5}, 0.3 * \dfrac{5}{0.5}, 0.4 * \dfrac{5}{0.5}, 0.5 \right.$<br>$\left. * \dfrac{5}{100}, 0.6 * \dfrac{5}{100}, 0.7 * \dfrac{5}{32}, 0.8 * \dfrac{5}{32}, 0.9 \right.$<br>$\left. * \dfrac{5}{100}, 1 * 5/100 \right]$ | 10 | [-5, 5] | 0 |

TABLE 6
"The 100-Digit Challenge:" CEC-C06 2019 Standards [31].

| No. | Functions | Dimension | Range | $f_{min}$ |
|-----|-----------|-----------|-------|-----------|
| 1 | Storn's Chebyshev Polynomial Fitting Problem | 9 | [-8192, 8192] | 1 |
| 2 | Inverse Hilbert Matrix Problem | 16 | [-16384, 16384] | 1 |
| 3 | Lennard-Jones Minimum Energy Cluster | 18 | [-4,4] | 1 |
| 4 | Rastrigin's Function | 10 | [-100, 100] | 1 |
| 5 | Griewangk's Function | 10 | [-100, 100] | 1 |
| 6 | Weierstrass Function | 10 | [-100, 100] | 1 |
| 7 | Modified Schwefel's Function | 10 | [-100, 100] | 1 |
| 8 | Expanded Schaffer's F6 Function | 10 | [-100, 100] | 1 |
| 9 | Happy Cat Function | 10 | [-100, 100] | 1 |
| 10 | Ackley Function | 10 | [-100, 100] | 1 |

NOTE: Readers who concern to know more information about CEC benchmarks can access this paper [31].